\documentclass[]{article}

\usepackage{amsmath}
\usepackage{balance}
\usepackage{booktabs}
\usepackage{graphicx}
\usepackage{multirow}
\usepackage{rotating}
\usepackage{soul}
\usepackage{url}
\usepackage[utf8]{inputenc}
\usepackage{amsfonts}
\usepackage{amssymb}
\usepackage{amsthm}
\usepackage[normalem]{ulem}
\usepackage{multicol}
\usepackage{hyperref}

\newif\ifdraft
\drafttrue

\newcommand{\FP}{\mathrm{FP}}
\newcommand{\FN}{\mathrm{FN}}

\begin{document}

\title{Cross-Lingual Sentiment Quantification}

\author{Andrea Esuli,
Alejandro Moreo,
Fabrizio Sebastiani\\
Istituto di Scienza e Tecnologie dell'Informazione \\
Consiglio Nazionale delle Ricerche\\ 56124 Pisa, Italy \\
Email:\{\textsf{firstname.lastname}\}@isti.cnr.it}

\date{}

\maketitle

\begin{abstract}
\noindent \emph{Sentiment Quantification} (i.e., the task of
estimating the relative frequency of sentiment-related classes ---
such as \textsf{Positive} and \textsf{Negative} --- in a set of
unlabelled documents) is an important topic in sentiment analysis, as
the study of sentiment-related quantities and trends across a
population is often of higher interest than the analysis of individual
instances.  In this work we propose a method for \emph{Cross-Lingual
Sentiment Quantification}, the task of performing sentiment
quantification when training documents are available for a source
language $\mathcal{S}$ but not for the target language $\mathcal{T}$
for which sentiment quantification needs to be
performed. Cross-lingual sentiment quantification (and cross-lingual
\emph{text} quantification in general) has never been discussed before
in the literature; we establish baseline results for the binary case
by combining state-of-the-art quantification methods with methods
capable of generating cross-lingual vectorial representations of the
source and target documents involved. We present experimental results
obtained on publicly available datasets for cross-lingual sentiment
classification; the results show that the presented methods can
perform cross-lingual sentiment quantification with a surprising level
of accuracy.
\end{abstract}


\section{Introduction}
\label{sec:introduction}

\noindent In \emph{Cross-Lingual Text Classification}, documents may
be expressed in either a \emph{source} language $\mathcal{S}$ or a
\emph{target} language $\mathcal{T}$, and training documents are
available only for $\mathcal{S}$ but not for $\mathcal{T}$;
cross-lingual text classification thus consists of leveraging the
training documents in the source language in order to train a
classifier for the target language, also using the fact that the
classification scheme $\mathcal{C}$ is the same for both $\mathcal{S}$
and $\mathcal{T}$. Cross-lingual text classification has been widely
investigated in the
literature~\cite{prettenhofer2011cross,Moreo:2016fg}. A companion task
which instead has never been tackled, and which is the object of this
paper, is \emph{Cross-Lingual Text Quantification}, the task of
performing ``quantification'' across a source language $\mathcal{S}$
and a target language $\mathcal{T}$. \emph{Quantification} is a
supervised learning task that consists of predicting, given a set of
classes $\mathcal{C}$ and a set $D$ (a \emph{sample}) of unlabelled
items drawn from some domain $\mathcal{D}$, the \emph{prevalence}
(i.e., relative frequency) $p_{c}(D)$ of each class $c\in\mathcal{C}$
in $D$. Put it another way, given an unknown distribution
$p_{\mathcal{C}}(D)$ of the members of $D$ across $\mathcal{C}$ (the
\emph{true distribution}), quantification consists in generating a
\emph{predicted distribution} $\hat{p}_{\mathcal{C}}(D)$ that
approximates $p_{\mathcal{C}}(D)$ as accurately as
possible~\cite{Gonzalez:2017it}.

Quantification is especially important for application fields
characterised by an interest in aggregate (rather than individual)
data, such as the social sciences, market research, political science,
and epidemiology. These disciplines often face the need to label data
in highly dynamic scenarios~\cite{Ebrahimi:2017jt}, i.e., scenarios in
which the distribution of data in the unlabelled set may be very
different from the distribution of data in the training set. In such
contexts, accurate class prevalence estimation may be challenging, due
to the fact that the ``iid assumption'' on which standard learning
methods are based (i.e., the assumption that the training set and the
test set are identically and independently sampled from the
\emph{same} data distribution) is obviously not verified.

This paper is about performing Cross-Lingual \emph{Sentiment}
Quantification. Sentiment quantification~\cite{Esuli:2010fk} is the
task of interest in all contexts in which the results of sentiment
analysis are to be analyzed at the aggregate level. For instance,
hardly anyone among those who perform sentiment analysis for Twitter
data are interested in determining the sentiment conveyed by a single
tweet; in most such applications, figuring out \emph{the percentage}
of tweets that exhibit a certain sentiment is the real goal, which
shows that quantification (and not classification) should be the task
to focus on~\cite{Gao:2016uq}. This paper adds cross-linguality to the
picture, thus addressing those application contexts characterized by
the absence of training data for the ``target'' language of interest,
and the presence of training data for a different ``source''
language. Everything we say in this paper straightforwardly extends to
dealing with the simultaneous presence of several source languages
and/or several target languages.

In principle, quantification can be straightforwardly solved via
classification, i.e., by training a classifier $h$ using training data
labelled according to $\mathcal{C}$, classifying the unlabelled data
in $D$ via $h$, and counting, for each $c\in\mathcal{C}$, how many
items in $D$ have been attributed to $c$ (the ``classify and count''
method). However, research has conclusively
shown~\cite{Barranquero:2015fr,Bella:2010kx,Esuli:2018rm,Forman:2008kx}
that this approach leads to suboptimal quantification accuracy. To see
this consider that a binary classifier $h_{1}$ for which $\FP=20$ and
$\FN=20$ ($\FP$ and $\FN$ standing for the ``false positives'' and
``false negatives'', respectively, that it has generated on a given
dataset) is worse, in terms of classification accuracy, than a
classifier $h_{2}$ for which, on the same dataset, $\FP=18$ and
$\FN=20$. However, $h_{1}$ is intuitively a better binary quantifier
than $h_{2}$; indeed, $h_{1}$ is a perfect quantifier, since $\FP$ and
$\FN$ are equal and thus, when it comes to class frequency estimation,
compensate each other, so that the distribution of the test items
across the class and its complement is estimated perfectly.  Since
classification and quantification pursue different goals,
quantification should be tackled as a task of its own, using different
evaluation measures and, as a result, different learning algorithms.

\noindent In this paper we establish baseline results for (binary)
cross-lingual sentiment quantification by combining a number of
quantification methods with state-of-the-art cross-lingual projection
methods, i.e., methods capable of generating language-agnostic
vectorial representations of the source and target documents involved.
For performing this latter task we explore \emph{Structural
Correspondence Learning} ($\mathrm{SCL}$~\cite{prettenhofer2011cross})
and \emph{Distributional Correspondence Indexing}
($\mathrm{DCI}$~\cite{Moreo:2016fg}), since (i) $\mathrm{SCL}$ is
arguably the most representative cross-lingual projection method in
the literature (and thus a mandatory baseline in lab experiments of
related research), while $\mathrm{DCI}$ is a cross-lingual projection
method that has recently demonstrated state-of-the-art performance in
cross-lingual text classification~\cite{Moreo:2018db}, and (ii) both
methods provide a general procedure for projecting source and target
documents onto a common vector space, and (iii) the code implementing
both methods is publicly available and easily modifiable.  Other
cross-lingual methods proposed in the literature learn representations
that are dependent on the set of unlabelled documents to classify (in
lab experiments: the test set). This implicitly means that each new
unlabelled set to quantify upon would require retraining from scratch,
something that would prove prohibitive in the experimental setting of
quantification (see Section \ref{sec:experiments}).

The rest of the paper is organized as
follows. Section~\ref{sec:method} describes the cross-lingual
sentiment quantification methods we use; Section~\ref{sec:experiments}
tests the presented methods on standard datasets for cross-lingual
sentiment classification, while Section~\ref{sec:conclusions}
concludes by discussing avenues for further research.


\section{Method}
\label{sec:method}

\noindent Different quantification methods have been proposed that
exploit the classification outcomes that a previously trained
classifier delivers on unlabelled data.  We explore different
cross-lingual sentiment quantification methods that result from the
combination of a cross-lingual projection method
(Section~\ref{sec:docrep}), a ``classify and count'' policy (Section
\ref{sec:classcount}), and an estimate correction method (Section
\ref{sec:correction}). In this paper we only address the binary case,
where the classes \{\textsf{Positive},\textsf{Negative}\} are
indicated as $\mathcal{C}=\{\oplus,\ominus\}$.


\subsection{Cross-Lingual Document Representations}
\label{sec:docrep}

\noindent In cross-lingual applications, $\mathrm{SCL}$ and
$\mathrm{DCI}$ rely on the concept of \emph{pivot term} (or simply
\emph{pivot})~\cite{Blitzer:2007gf} in order to bridge the gap between
the different feature spaces which the different languages
generate. In such contexts, pivots are defined as highly predictive
pairs of translation-equivalent terms which behave in a similar way in
their respective languages. Typical examples of pivots for
sentiment-related applications are adjectives with domain-independent
meaning such as ``excellent'' or ``poor'', and partially
domain-dependent terms such as ``fancy'' (as found, e.g., in the arts
and crafts domain and in the clothing domain) or ``masterpiece'' (as
found, e.g., in the book domain, movie domain, and music domain), with
their respective translations in other languages.

A common strategy to select the pivots automatically consists of
taking the top elements from a list of terms ranked according to their
mutual information to the label representing the domain (as computed
from source-language training data), and filtering out those
candidates whose translation equivalent shows a substantial prevalence
drift in the target language. A word translation oracle, with a fixed
budget of allowed calls, is assumed available.

Once pivots are selected, different methods can be defined in order to
produce cross-lingual vectorial representations. Both $\mathrm{SCL}$
and $\mathrm{DCI}$ first represent documents as vectors $\mathbf{x}$
in a (weighted) bag-of-words model of dimension $|V|$ (with $V$ being
the vocabulary), and then apply a linear projection (parameterized by
a matrix $\theta\in\mathbb{R}^{|V|L}$) of type
$\mathbf{x}^\top\theta$, thus mapping $|V|$-dimensional vectors into
$L$-dimensional vectors in a cross-lingual latent space. To achieve
this, the unlabelled collections from the source and target domains
are inspected. Once defined, the matrix can be subsequently used to
project source documents (to train a classifier) and target documents
(to classify them).

$\mathrm{SCL}$ builds the projection matrix by resolving an auxiliary
prediction problem for each pair of translation-equivalent pivot
terms.  Each problem consists of predicting the presence of a pivot
term based on the observation of the other terms. By solving the
auxiliary problems (via linear classification), structural
correspondences among terms and pivots are captured and collected as a
matrix of correlations. This matrix is later decomposed using
truncated SVD to generate the final projection matrix $\theta$.
$\mathrm{DCI}$ relies instead on the distributional hypothesis to
directly model correspondences between terms and pivots.  Each row of
the projection matrix $\mathrm{DCI}$ computes represents a term
profile, where each dimension quantifies the degree of correspondence
(as measured by a \emph{distributional correspondence function}) of
the term to a pivot.


\subsection{Classifying and Counting}
\label{sec:classcount}

\noindent An obvious way to solve quantification is by aggregating the
scores assigned by a classifier to the unlabelled documents.

In connection to each of $\mathrm{SCL}$ and $\mathrm{DCI}$ we
experiment with two different aggregation methods, one that uses a
``hard'' classifier (i.e., a classifier
$h_{\oplus}:\mathcal{D}\rightarrow \{0,1\}$ that outputs binary
decisions, 0 for $\ominus$ and 1 for $\oplus$) and one that uses a
``soft'' classifier (i.e., a classifier
$s_{\oplus}:\mathcal{D}\rightarrow [0,1]$ that outputs posterior
probabilities $\Pr(\oplus|\mathbf{x})$, representing the probability
that the classifier attributes to the fact that $\mathbf{x}$ belongs
to the $\oplus$ class). Of course,
$\Pr(\ominus|\mathbf{x})=(1-\Pr(\oplus|\mathbf{x}))$.

The (trivial) \emph{classify and count} ($\mathrm{CC}$) quantifier
then comes down to computing
\begin{align}
  \label{eq:cc}
  \hat{p}_{\oplus}^{\mathrm{CC}}(D) & = \frac{\sum_{\mathbf{x}\in D}
                                      h_{\oplus}(\mathbf{x})}{|D|}
\end{align}
\noindent while the \emph{probabilistic classify and count} quantifier
($\mathrm{PCC}$~\cite{Bella:2010kx}) is defined by
\begin{align}
  \label{eq:pcc}
  \hat{p}_{\oplus}^{\mathrm{PCC}}(D) & = \frac{\sum_{\mathbf{x}\in D}
                                       s_{\oplus}(\mathbf{x})}{|D|}
\end{align}

\noindent Of course, for any method $M$ we have
$\hat{p}_{\ominus}^{\mathit{M}}(D)=(1-\hat{p}_{\oplus}^{\mathit{M}}(D))$.


\subsection{Adjusting the Results of Classify and Count}
\label{sec:correction}

\noindent A popular quantification method consists of applying an
\emph{adjustment} to the prevalence $\hat{p}_{\oplus}(D)$ estimated
via ``classify and count''. It is easy to check that, in the binary
case, the true prevalence $p_{\oplus}(D)$ and the estimated prevalence
$\hat{p}_{\oplus}(D)$ are such that
\begin{equation}
  \label{eq:exactacc} 
  p_{\oplus}(D) = \frac{\hat{p}_{\oplus}^{\mathrm{CC}}(D) - \mathit{fpr}_{h}}{\mathit{tpr}_{h} - \mathit{fpr}_{h}}
\end{equation}
\noindent where $\mathit{tpr}_{h}$ and $\mathit{fpr}_{h}$ stand for
the \emph{true positive rate} and \emph{false positive rate} of the
classifier $h_{\oplus}$ used to obtain
$\hat{p}_{\oplus}^{\mathrm{CC}}$.  The values of $\mathit{tpr}_{h}$
and $\mathit{fpr}_{h}$ are unknown, but can be estimated via $k$-fold
cross-validation on the training data.  In the binary case this comes
down to using the results $h_{\oplus}(\mathbf{x})$ obtained in the
$k$-fold cross-validation (i.e., $\mathbf{x}$ ranges on the training
documents) in equations
\begin{align}
  \begin{split}
    \label{eq:tprandfpr}
    \hat{\mathit{tpr}_{h}} = \frac{\sum_{\mathbf{x}\in
    \oplus}h_{\oplus}(\mathbf{x})}{|\{\mathbf{x}\in \oplus\}|}
    \hspace{3em} \hat{\mathit{fpr}_{h}} =\frac{\sum_{\mathbf{x}\in
    \ominus}h_{\oplus}(\mathbf{x})}{|\{\mathbf{x}\in \ominus\}|}
  \end{split}
\end{align}
\noindent We obtain estimates of $p_{\oplus}^{\mathrm{ACC}}(D)$, which
define the \emph{adjusted classify and count}
method~\cite{Forman:2008kx} ($\mathrm{ACC}$) by replacing
$\mathit{tpr}_{h}$ and $\mathit{fpr}_{h}$ in Equation
\ref{eq:exactacc} with the estimates of Equation \ref{eq:tprandfpr},
i.e.,
\begin{equation}
  \label{eq:acc} 
  \hat{p}_{\oplus}^{\mathrm{ACC}}(D) = \frac{\hat{p}_{\oplus}^{\mathrm{CC}}(D) - \hat{\mathit{fpr}_{h}}}{\hat{\mathit{tpr}_{h}} - \hat{\mathit{fpr}_{h}}}
\end{equation}
\noindent If the soft classifier $s_{\oplus}(\mathbf{x})$ is used in
place of $h_{\oplus}(\mathbf{x})$, analogues of
$\hat{\mathit{tpr}_{h}}$ and $\hat{\mathit{fpr}_{h}}$ from Equation
\ref{eq:tprandfpr} can be defined as
\begin{align}
  \begin{split}\label{eq:tprandfpr2}
    \hat{\mathit{tpr}_{s}} = \frac{\sum_{\mathbf{x}\in
    \oplus}s_{\oplus}(\mathbf{x})}{|\{\mathbf{x}\in \oplus\}|}
    \hspace{3em} \hat{\mathit{fpr}_{s}} =\frac{\sum_{\mathbf{x}\in
    \ominus}s_{\oplus}(\mathbf{x})}{|\{\mathbf{x}\in \ominus\}|}
  \end{split}
\end{align}
\noindent We obtain $p_{\oplus}^{\mathrm{PACC}}(D)$ estimates, which
define the \emph{probabilistic adjusted classify and count} method
($\mathrm{PACC}$~\cite{Bella:2010kx}), by replacing all factors in the
right-hand side of Equation \ref{eq:acc} with their ``soft''
counterparts from Equations \ref{eq:pcc} and \ref{eq:tprandfpr2},
i.e.,
\begin{equation}
  \label{eq:pacc} 
  \hat{p}_{\oplus}^{\mathrm{PACC}}(D) = \frac{\hat{p}_{\oplus}^{\mathrm{PCC}}(D) - \hat{\mathit{fpr}_{s}}}{\hat{\mathit{tpr}_{s}} - \hat{\mathit{fpr}_{s}}}
\end{equation}
\noindent $\mathrm{ACC}$ and $\mathrm{PACC}$ define two simple linear
adjustments to the aggregated scores of general-purpose classifiers.
We also investigate the use of a more recently proposed adjustment
method beased on deep learning, called QuaNet~\cite{Esuli:2018rm}.
QuaNet models a neural \emph{non-linear} adjustment by taking as input
all estimated prevalences from Equations \ref{eq:cc}, \ref{eq:pcc},
\ref{eq:acc}, \ref{eq:pacc} (i.e., $\hat{p}_{\oplus}^{\mathrm{CC}}$,
$\hat{p}_{\oplus}^{\mathrm{ACC}}$, $\hat{p}_{\oplus}^{\mathrm{PCC}}$,
$\hat{p}_{\oplus}^{\mathrm{PACC}}$), several statistics (the
$\hat{\mathit{tpr}}_{h}$, $\hat{\mathit{fpr}}_{h}$,
$\hat{\mathit{tpr}}_{s}$, $\hat{\mathit{fpr}}_{s}$ estimates from
Equations \ref{eq:tprandfpr} and \ref{eq:tprandfpr2}), the posterior
probabilities $\Pr(\oplus|\mathbf{x})$ for each document $\mathbf{x}$,
and the document vectors themselves.  QuaNet relies on a recurrent
neural network to produce ``quantification embeddings'' (i.e., dense,
multi-dimensional representations of the information relevant to
quantification observed from the input data), which are then used to
generate the final prevalence estimates.


\section{Experiments}
\label{sec:experiments}

\noindent In this section we report on the experiments we have run on
sentiment classification data in order to empirically evaluate the
effectiveness of our cross-lingual sentiment quantification
approaches.  We test each of the $2\times 5=10$ combinations resulting
from 2 approaches to generating cross-lingual projections
($\mathrm{SCL}$ and $\mathrm{DCI}$) and 5 approaches to performing
quantification ($\mathrm{CC}$, $\mathrm{PCC}$, $\mathrm{ACC}$,
$\mathrm{PACC}$, and Quanet). The code to replicate all these
experiments is available from GitHub~\cite{clquant}. Note that a
dataset for sentiment classification is also a dataset for sentiment
quantification, since from the manually assigned labels of the test
data one can compute the true class prevalences $p_{\oplus}(D)$ and
$p_{\ominus}(D)$ by simply counting.


\subsection{System setup}
\label{sec:systemsetup}

\noindent We use the \textsc{Nut} package~\cite{nut} for $\mathrm{SCL}$
and the \textsc{PyDCI} package~\cite{Moreo:2018db} for $\mathrm{DCI}$
in order to generate the vectorial representations of all training and
test documents.  As the hard classifiers, we stick to the ones used by
the original proponents of $\mathrm{SCL}$ and $\mathrm{DCI}$, i.e., a
linear classifier trained via Elastic Net~\cite{Zou:2005tk}
(implemented via the \textsc{Bolt} package~\cite{bolt}) for
$\mathrm{SCL}$, and a linear classifier trained via SVMs (implemented
via the \textsc{Scikit-Learn} package~\cite{Pedregosa:2011yo}) for
$\mathrm{DCI}$.  As the soft classifier we instead use one trained via
logistic regression (in its \textsc{Scikit-Learn} implementation) for
both $\mathrm{SCL}$ and $\mathrm{DCI}$, since such classifiers are
known to return ``well-calibrated'' posterior probabilities.

The last point is fundamental for Equations \ref{eq:pcc},
\ref{eq:tprandfpr2}, \ref{eq:pacc} to return accurate values, since
``well calibrated probabilities'' is essentially a synonym of
``good-quality probabilities''.  Posterior probabilities
$\Pr(c|\mathbf{x})$ are said to be \emph{well calibrated} when, given
a sample $D$ drawn from some population,
$$\lim_{|D|\rightarrow \infty}\frac{|\{\mathbf{x}\in c|
\Pr(c|\mathbf{x})=\alpha\}|}{|\{\mathbf{x}\in D|
\Pr(c|\mathbf{x})=\alpha\}|}=\alpha.$$ Intuitively, this property
implies that, as the size of the sample $D$ goes to infinity, e.g.,
90\% of the documents $\mathbf{x}\in D$ that are assigned a well
calibrated posterior probability $\Pr(c|\mathbf{x})=0.9$ belong to
class $c$.  Some classifiers (e.g., those trained via logistic
regression~\cite{Zadrozny:2002eu}) are known to return well calibrated
probabilities. The posterior probabilities returned by some other
classifiers (e.g., those trained via na\"ive Bayesian methods~\cite{Domingos:1997dl}) are known instead to be not well
calibrated. Yet some other classifiers (e.g., those trained via SVMs)
do not return posterior probabilities, but generic confidence
scores. In these two last cases it is possible to map the obtained
posterior probabilities / confidence scores into well calibrated
posterior probabilities by means of some ``calibration'' method~\cite{Platt:2000fk,Zadrozny:2002eu}.

We set all the hyper-parameters in $\mathrm{SCL}$ (number $m$ of
pivots, minimum support frequency $\phi$ for pivot candidates,
dimensionality $k$ of the cross-lingual representation, and the
Elastic Net coefficient $\alpha$) to ($m=450$, $\phi=30$, $k=100$,
$\alpha=0.85$), i.e., to the values found optimal in previous
literature~\cite{prettenhofer2011cross} when optimizing for the German
book review task. Along with previous work~\cite{Moreo:2018db}, in
$\mathrm{DCI}$ we set the number of pivots and minimum support to
$m=450$ and $\phi=30$.  The dimensionality is $k=450$ by definition,
since in $\mathrm{DCI}$ each pivot corresponds to a dimension.  In
preliminary experiments we had used the same value $k=450$ both for
$\mathrm{DCI}$ and $\mathrm{SCL}$, on grounds of ``fairness''.  The
results for $\mathrm{SCL}$ were slightly worse with respect to using
$k=100$; for $\mathrm{SCL}$ we thus decided to stick to the $k=100$
value originally used by the creators of
$\mathrm{SCL}$~\cite{prettenhofer2011cross}.  As the distributional
correspondence function we use cosine, since it is the one which
delivered the best performance in previously published
experiments~\cite{Moreo:2018db}. For each setup we independently
optimize the parameter $C$ (which controls the regularization strength
in the SVM and in the logistic regressor) via grid search in the log
space defined by $C\in\{10^i\}_{i=-5}^{5}$, and via 5-fold
cross-validation.  The classifiers with the optimized hyper-parameters
are then used in a 10-fold cross-validation run on the training data
to produce the $\hat{\mathit{tpr}_{h}}$ and $\hat{\mathit{fpr}_{h}}$
estimates.

For the neural correction of QuaNet we use its publicly available
implementation linked from the original paper~\cite{quanet}.  We
optimize the hyper-parameters of QuaNet using the German book review
task (as done by Prettenhofer and Stein~\cite{prettenhofer2011cross});
we end up using 64 hidden units in the recurrent cell of a two-layer
stacked bidirectional LSTM, 1024 and 512 hidden units in the
next-to-last feed-forward layers, and a drop probability of 0.  We set
the rest of the parameters to the same values as in the original
QuaNet paper~\cite{Esuli:2018rm}.


\subsection{Experimental setting}
\label{sec:experimentalsetting}

\noindent We use the Webis-CLS-10 dataset~\cite{prettenhofer2011cross,
clsacl10} as the benchmark for our experiments.  Webis-CLS-10 is a
dataset originally proposed for cross-lingual sentiment classification
experiments, and consisting of Amazon product reviews written in four
languages (English, German, French, Japanese) and concerning three
product domains (Books, DVDs, Music).  There are 2,000 training
documents, 2,000 test documents, and a number of unlabelled documents
ranging from 9,000 to 50,000 for each combination of language and
domain.  The examples of $\oplus$ and $\ominus$ (which indicate
positive and negative sentiment, resp.) are perfectly balanced (i.e.,
50\% each) in all sets (training, test, unlabelled). Following a
consolidated practice in cross-lingual text classification, we always
use English as the source language.  We use the publicly available
pre-processed version of the dataset~\cite{prettenhofer2011cross,
clsacl10}, where terms correspond to uni-grams.

As the measures of quantification error we use \emph{Absolute Error}
($\mathrm{AE}$), \emph{Relative Absolute Error} ($\mathrm{RAE}$), and
the \emph{Kullback-Leibler Divergence} ($\mathrm{KLD}$), defined as:
\begin{align}
  \mathrm{AE}(p,\hat{p},D) & =\frac{1}{|\mathcal{C}|}\sum_{c\in \mathcal{C}}|\hat{p}_{c}(D)-p_{c}(D)| 
                             \label{eq:ae} \\
  \mathrm{RAE}(p,\hat{p},D) & =\frac{1}{|\mathcal{C}|}\sum_{c\in 
                              \mathcal{C}}\displaystyle\frac{|\hat{p}_{c}(D)-p_{c}(D)|}{p_{c}(D)} 
                              \label{eq:rae} \\
  \mathrm{KLD}(p,\hat{p},D) & =\sum_{c\in \mathcal{C}} 
                              p_{c}(D)\log\frac{p_{c}(D)}{\hat{p}_{c}(D)} 
                              \label{eq:kld} 
\end{align}
\noindent since they are the most frequently used measures for
evaluating quantification error~\cite{Sebastiani:2020qf}.

The evaluation of a quantifier cannot be carried out on the basis on
one single set of test documents. The reason is that, while in text
classification experiments a test set consisting of $n$ documents
enables the evaluation of $n$ different decision outcomes, in
quantification the same test set would only allow to validate one
single prevalence prediction. In order to allow statistically
significant comparisons, Forman~\cite{Forman:2008kx} proposed to run
quantification experiments on a set of test samples, randomly sampled
from the original set of test documents at different prevalence
levels.  Along with Forman~\cite{Forman:2008kx}, as the range of
prevalences for the $\oplus$ class we use \{0.01, 0.05, 0.10, \ldots,
0.90, 0.95, 0.99\}.
Similarly to previous work~\cite{Esuli:2018rm}, we generate 100 random
samples for each of the 21 prevalence levels, and report
quantification error as the average across $21\times 100 = 2100$ test
samples.
All samples consist of 200 documents.  For each target language
(German, French, Japanese) and product domain (Books, DVD, Music) the
samples are the same across the different methods, which will enable
us to evaluate the statistical significance of the differences in
performance; to this aim, we rely on the non-parametric Wilcoxon
signed-rank test on paired samples.

For each combination of target language and product domain,
Table~\ref{tab:results} reports quantification error (for each CLTQ
method and for each evaluation measure) as an average across the 2100
test samples; we recall that English is always used as the source
language, so that, e.g., the ``German Books'' experiment is about
training on English book reviews and testing on German book reviews.
Since QuaNet depends on a stochastic optimization,
Table~\ref{tab:results} reports the average and standard deviation
across 10 runs.

\begin{table}[t]
  \caption{Cross-lingual sentiment quantification results for
  Webis-CLS-10. \textbf{Boldface} indicates the best
  result. Superscripts $\dag$ and $\dag\dag$ denote the method (if
  any) whose score is not statistically significantly different from
  the best one at $\alpha=0.05$ ($\dag$) or at $\alpha=0.005$
  ($\dag\dag$).}
  \label{tab:results}
  \resizebox{\textwidth}{!}{
  \begin{tabular}{|c|cc|lllll|lllll|}
    \hline
    & \multicolumn{1}{c}{Target}   & & \multicolumn{5}{c}{$\mathrm{SCL}$} & \multicolumn{5}{|c|}{$\mathrm{DCI}$} \\
    & \multicolumn{1}{c}{Language} & Domain &     $\mathrm{CC}$ &            $\mathrm{ACC}$ &    $\mathrm{PCC}$ &   $\mathrm{PACC}$ &              QuaNet &     $\mathrm{CC}$ &             $\mathrm{ACC}$ &    $\mathrm{PCC}$ &            $\mathrm{PACC}$ &                       QuaNet \\
    \hline  
    \multirow{10}{*}{\begin{sideways}$\mathrm{AE}$\end{sideways}}
    & German & Books   &  0.092 &          0.040 &  0.237 &  0.375 &  0.203 ($\pm$0.006) &  0.090 &           0.037 &  0.119 &  \textbf{0.027} &           0.030 ($\pm$0.002) \\
    &    German & DVDs     &  0.104 &          0.045 &  0.221 &  0.331 &  0.178 ($\pm$0.009) &  0.086 &           0.030 &  0.147 &  \textbf{0.028} &   0.030 ($\pm$0.003)\dag\dag \\
    &    German & Music   &  0.097 &  0.037\dag\dag &  0.151 &  0.101 &  0.072 ($\pm$0.007) &  0.078 &   0.037\dag\dag &  0.109 &   0.039\dag\dag &  \textbf{0.030} ($\pm$0.002) \\
    &    French & Books   &  0.098 &          0.037 &  0.202 &  0.288 &  0.151 ($\pm$0.007) &  0.098 &           0.038 &  0.122 &  \textbf{0.025} &           0.036 ($\pm$0.003) \\
    &    French & DVDs     &  0.110 &          0.056 &  0.174 &  0.113 &  0.072 ($\pm$0.002) &  0.091 &           0.037 &  0.117 &  \textbf{0.027} &           0.045 ($\pm$0.005) \\
    &    French & Music   &  0.119 &          0.060 &  0.178 &  0.090 &  0.072 ($\pm$0.001) &  0.074 &           0.030 &  0.160 &  \textbf{0.024} &           0.047 ($\pm$0.010) \\
    &    Japanese & Books &  0.127 &          0.072 &  0.194 &  0.124 &  0.095 ($\pm$0.002) &  0.117 &  \textbf{0.060} &  0.174 &           0.064 &           0.073 ($\pm$0.003) \\
    &    Japanese & DVDs   &  0.131 &          0.079 &  0.329 &  0.485 &  0.270 ($\pm$0.005) &  0.104 &           0.045 &  0.128 &  \textbf{0.037} &           0.058 ($\pm$0.006) \\
    &    Japanese & Music &  0.118 &          0.059 &  0.242 &  0.377 &  0.228 ($\pm$0.007) &  0.092 &           0.029 &  0.161 &  \textbf{0.027} &           0.044 ($\pm$0.009) \\
    \cline{2-13}
    &    \multicolumn{2}{c|}{Average}        &  0.111 &          0.054 &  0.214 &  0.254 &               0.149 &  0.092 &           0.038 &  0.138 &  \textbf{0.033} &                        0.044 \\
    \hline  
    \hline
    \multirow{10}{*}{\begin{sideways}$\mathrm{RAE}$\end{sideways}}
    &    German & Books   &  0.888 &      0.164 &  0.878 &  0.807 &  0.513 ($\pm$0.015) &  1.135 &           0.246 &  1.411 &  \textbf{0.136} &          0.248 ($\pm$0.034) \\
    &    German & DVDs     &  1.086 &      0.267 &  1.047 &  0.733 &  0.428 ($\pm$0.031) &  1.070 &           0.223 &  1.709 &  \textbf{0.144} &  0.234 ($\pm$0.020)\dag\dag \\
    &    German & Music   &  1.056 &  0.194\dag &  1.364 &  0.268 &  0.216 ($\pm$0.011) &  0.947 &   0.194\dag\dag &  1.310 &  \textbf{0.153} &  0.245 ($\pm$0.022)\dag\dag \\
    &    French & Books   &  1.021 &      0.313 &  1.041 &  0.666 &  0.383 ($\pm$0.025) &  1.227 &           0.407 &  1.426 &  \textbf{0.159} &          0.330 ($\pm$0.026) \\
    &    French & DVDs     &  1.307 &      0.682 &  1.642 &  0.475 &  0.543 ($\pm$0.019) &  0.938 &           0.176 &  1.284 &  \textbf{0.144} &          0.223 ($\pm$0.016) \\
    &    French & Music   &  1.310 &      0.496 &  2.099 &  1.181 &  0.817 ($\pm$0.026) &  0.834 &  \textbf{0.138} &  1.803 &           0.208 &      0.276 ($\pm$0.039)\dag \\
    &    Japanese & Books &  1.423 &      0.781 &  2.287 &  1.572 &  1.122 ($\pm$0.026) &  1.196 &  \textbf{0.450} &  1.935 &           0.639 &          0.570 ($\pm$0.032) \\
    &    Japanese & DVDs   &  1.392 &      0.785 &  0.833 &  0.947 &  0.557 ($\pm$0.012) &  1.097 &           0.292 &  1.380 &  \textbf{0.213} &          0.350 ($\pm$0.021) \\
    &    Japanese & Music &  1.232 &      0.304 &  0.910 &  0.806 &  0.527 ($\pm$0.016) &  0.973 &  \textbf{0.175} &  1.800 &       0.198\dag &          0.293 ($\pm$0.034) \\
    \cline{2-13}
    &        \multicolumn{2}{c|}{Average}        &  1.191 &      0.443 &  1.345 &  0.828 &               0.567 &  1.046 &           0.256 &  1.562 &  \textbf{0.222} &                       0.308 \\
    \hline  
    \hline
    \multirow{10}{*}{\begin{sideways}$\mathrm{KLD}$\end{sideways}}
    &        German & Books   &  0.041 &          0.016 &  0.194 &  1.778 &  0.274 ($\pm$0.043) &  0.040 &           0.032 &  0.062 &           0.028 &  \textbf{0.007} ($\pm$0.001) \\
    &    German & DVDs     &  0.050 &          0.013 &  0.172 &  0.987 &  0.139 ($\pm$0.034) &  0.038 &           0.019 &  0.086 &           0.028 &  \textbf{0.007} ($\pm$0.001) \\
    &    German & Music   &  0.045 &  0.017\dag\dag &  0.090 &  0.062 &  0.027 ($\pm$0.005) &  0.032 &           0.046 &  0.054 &           0.072 &  \textbf{0.008} ($\pm$0.001) \\
    &    French & Books   &  0.046 &  0.010\dag\dag &  0.146 &  0.748 &  0.115 ($\pm$0.024) &  0.046 &           0.014 &  0.064 &           0.014 &  \textbf{0.010} ($\pm$0.001) \\
    &    French & DVDs     &  0.055 &          0.019 &  0.111 &  0.055 &  0.029 ($\pm$0.001) &  0.040 &           0.012 &  0.060 &  \textbf{0.008} &           0.012 ($\pm$0.002) \\
    &    French & Music   &  0.062 &          0.021 &  0.114 &  0.040 &  0.028 ($\pm$0.000) &  0.030 &           0.040 &  0.097 &  \textbf{0.007} &           0.014 ($\pm$0.004) \\
    &    Japanese & Books &  0.068 &          0.028 &  0.132 &  0.065 &  0.043 ($\pm$0.001) &  0.060 &  \textbf{0.020} &  0.110 &           0.024 &           0.029 ($\pm$0.002) \\
    &    Japanese & DVDs   &  0.071 &          0.033 &  0.376 &  5.133 &  0.250 ($\pm$0.013) &  0.051 &           0.014 &  0.069 &  \textbf{0.011} &           0.020 ($\pm$0.003) \\
    &    Japanese & Music &  0.061 &          0.022 &  0.202 &  1.629 &  0.234 ($\pm$0.024) &  0.042 &           0.011 &  0.098 &  \textbf{0.009} &           0.013 ($\pm$0.004) \\
    \cline{2-13}
    &        \multicolumn{2}{c|}{Average}        &  0.055 &          0.020 &  0.171 &  1.166 &               0.127 &  0.042 &           0.023 &  0.078 &           0.022 &               \textbf{0.013} \\

    \hline
  \end{tabular}
  }
\end{table}


\subsection{Results}
\label{sec:results}

\noindent Overall, the results indicate that the combination
$\mathrm{DCI}$+$\mathrm{PACC}$ is the best performer in terms of
$\mathrm{AE}$ and $\mathrm{RAE}$, while $\mathrm{DCI}$+QuaNet seems to
behave slightly better in terms of $\mathrm{KLD}$. Given recent
theoretical results on the properties of evaluation measures for
quantification~\cite{Sebastiani:2020qf}, that indicate that
$\mathrm{AE}$ and $\mathrm{RAE}$ are to be preferred to
$\mathrm{KLD}$, this leads us to prefer
$\mathrm{DCI}$+$\mathrm{PACC}$.

If we look at the results in more detail, one aspect that emerges is
the substantial superiority of $\mathrm{DCI}$ over $\mathrm{SCL}$, as
witnessed by the fact that, for each combination of evaluation
measure, target language, and domain, the best performer always uses
$\mathrm{DCI}$ and not $\mathrm{SCL}$. This confirms previous
results~\cite{Moreo:2016fg} that showed the superiority of
$\mathrm{DCI}$ over $\mathrm{SCL}$ in monolingual sentiment
classification contexts.

In both $\mathrm{SCL}$ and $\mathrm{DCI}$ the ``hard'' classifier
tends to work comparatively better than the ``soft'' logistic
regressor, as indicated by the fact that $\mathrm{CC}$ tends to
outperform $\mathrm{PCC}$ and $\mathrm{ACC}$ tends (with some
exceptions) to outperform $\mathrm{PACC}$. As expected, $\mathrm{ACC}$
(the ``adjusted'' version of $\mathrm{CC}$) performs substantially
better than $\mathrm{CC}$ in all cases. What comes as a surprise,
though, is the fact that the remarkable benefit $\mathrm{PACC}$ brings
about in $\mathrm{DCI}$ with respect to its unadjusted variant
$\mathrm{PCC}$, is not consistently mirrored in the case of
$\mathrm{SCL}$ (where the effect of adjusting is instead harmful, and
especially so in terms of $\mathrm{KLD}$).

The neural, non-linear adjustment of QuaNet, when applied to
$\mathrm{DCI}$ vectors, performs somehow similarly to the best
performer in several cases, and actually delivers the lowest average
$\mathrm{KLD}$ error. That QuaNet does not perform as well with
$\mathrm{SCL}$ can be explained by two facts (which are not
independent of each other), i.e., the importance of the estimated
posterior probabilities within QuaNet, and the suboptimal ability (as
shown by the $\mathrm{PCC}$ and $\mathrm{PACC}$ results) in delivering
accurate posterior probabilities for $\mathrm{SCL}$ vectors that the
logistic regressor has shown.


\section{Conclusions}
\label{sec:conclusions}

\noindent 
The experiments we have performed show that structural correspondence
learning ($\mathrm{SCL}$) and distributional correspondence indexing
($\mathrm{DCI}$), two previously proposed methods for cross-lingual
text classification, can effectively be used in cross-lingual text
quantification, a task that had never been tackled before in the
literature. The tested methods yield quantification predictions that
are fairly close to the true prevalence; in terms of absolute error
(arguably the most easy-to-interpret error criterion), and on average,
the class prevalences predicted by $\mathrm{DCI}$+$\mathrm{PACC}$
differ from the true prevalences by a margin of 3.3\% on average,
while this difference is 5.4\% for
$\mathrm{SCL}$+$\mathrm{ACC}$. These results are encouraging,
especially if we consider the fact that the quantifier is trained on a
language different from the one on which quantification is performed
(for which no training data are assumed to exist), and that a range of
true prevalences different (and even extremely different) from the
ones of the training set are tested upon.

Note also that these results are a further confirmation of the fact
that, when our interest in automatically labelled data is at the
aggregate level only (and not at the individual level), using ``real''
quantification methods (instead of standard classification methods in
a ``classify and count'' fashion) is the way to go. To witness, in
terms of absolute error the use of $\mathrm{DCI}$+$\mathrm{PACC}$
allows to cut down quantification error to 3.3\% on average, a
substantial improvement with respect to the 9.2\% on average obtained
by just using $\mathrm{DCI}$ with a ``classify and count'' approach.

The combination of transfer learning (of which cross-lingual transfer
is an instance) with quantification is an interesting task in general,
that should prompt a body of dedicated research. We believe end-to-end
approaches for cross-lingual quantification, not necessarily relying
on classification as an intermediate step, would be worth exploring.
Likewise, a natural extension of this work would be to explore
applications of transfer learning to sentiment quantification
different from the cross-lingual one, such as cross-domain sentiment
quantification. Note also that, while this paper concentrates on a
very narrow aspect of sentiment analysis (namely,
\textsf{Positive}-\textsf{Negative} polarity detection), approaches
such as the ones championed here can be in principle extended to deal
with other labelling tasks in affective computing and sentiment
analysis~\cite{Cambria:2016cr}, such as finer-grained polarity
detection (e.g., using ordinal scales~\cite{DaSanMartino:2016jk}) or
joint topic-sentiment detection~\cite{Yang:2019to}.


\bibliographystyle{ieeeCSBib} \bibliography{Fabrizio,Alejandro,Andrea}

\end{document}